\pdfoutput=1

\documentclass[11pt]{article}

\usepackage[final]{acl}

\usepackage{times}
\usepackage{latexsym}

\usepackage[T1]{fontenc}

\usepackage[utf8]{inputenc}

\usepackage{microtype}

\usepackage{inconsolata}

\usepackage{graphicx}

\usepackage{amsmath}
\usepackage{amssymb}

\usepackage{pifont}  

\newcommand{\cmark}{\ding{51}}  
\newcommand{\xmark}{\ding{55}}  
\newcommand{\warn}{\ding{69}}

\title{Enhancing BERT Fine-Tuning for Sentiment Analysis in Lower-Resourced Languages}

\author{Jozef Kubík \and Marek Šuppa \and Martin Takáč \\
         Faculty of Mathematics Physics and Informatics \\ Comenius University in Bratislava, Slovakia}

\begin{document}
\maketitle
\begin{abstract}
Limited data for low-resource languages typically yield weaker language models (LMs). Since pre-training is compute-intensive, it is more pragmatic to target improvements during fine-tuning. In this work, we examine the use of Active Learning (AL) methods augmented by structured data selection strategies which we term 'Active Learning schedulers,' to boost the fine-tuning process with a limited amount of training data. We connect the AL to data clustering and propose an integrated fine-tuning pipeline that systematically combines AL, clustering, and dynamic data selection schedulers to enhance model's performance. Experiments in the Slovak, Maltese, Icelandic and Turkish languages show that the use of clustering during the fine-tuning phase together with AL scheduling can simultaneously produce annotation savings up to 30\% and performance improvements up to four F1 score points, while also providing better fine-tuning stability.
\end{abstract}

\section{Introduction}

LMs such as BERT \citep{devlin2018bert} are considered breakthrough models in many areas. Although nowadays most of the research is done using prompting techniques on large foundational models \citep{shin2023prompt}, BERT family models are still used in lower-resource scenarios for a number of reasons. They allow full control\footnote{By full control we mean direct control over hyperparameters, architectural extensions, and training regimes.}  over fine-tuning, and their size and hardware requirements make them more easily feasible to use in the research and real-world environments. In contrast, using commercial models at scale can be expensive and often relies on Internet connection, meaning their use is not possible when the data in question cannot be shared with third parties (e.g. for privacy reasons).

There is a huge performance gap between high-resource languages, such as English, and lower-resourced languages in both prompting models and BERT-like models. However, this gap can be closed with clever modifications in both training phases, with the BERT family models being easier to train than GPT-like models \citep{achiam2023gpt} due to their size and data requirements.

BERT's effectiveness often depends on large annotated datasets, highlighting the need for more efficient data exploration under limited annotation budgets. To address this, we propose a novel fine-tuning pipeline for lower-resourced language BERT models on the classification task. Our languages for the experiments representing lower-resourced languages will be Maltese, Icelandic, and Slovak, the last being the most popular of them spoken by approximately 5 million people \citep{short2018czech}. To show possible generalization on more popular languages not considered low-resource but rather medium- to high-resource, we also conduct experiments on the Turkish language.

\section{Related Work}

The definition of low-resource can be interpreted in different ways depending on the language or setting \citep{nigatu2024zeno}. For the purpose of this study, we define low-resource operationally as a setting with severely limited annotated data.

Popular techniques in low-resource scenarios include adapting smaller models trained in English to other languages by modifying parts of the architecture \citep{al2017aroma} or transferring the knowledge of models pre-trained in high-resource languages to other models fine-tunable in low-resource languages \citep{heinzerling2017bpemb,bojanowski2017enriching}. One of the alternative approaches, where the model itself helps identify the data samples which will help it to learn more efficiently, is called Active learning \citep{settles2009active}. Popular AL methods for classification tasks are pool-based methods using uncertainty sampling \citep{lewis1995sequential} such as entropy \citep{shannon1948mathematical} sampling. Fine-tuning with Active learning can also be enhanced by using adapter modules \citep{jukic2023parameter}, Reinforcement Learning \citep{wertz2023reinforced}, or Epistemic Neural networks (ENNs) \citep{osband2022fine}. Some methods  (especially when fine-tuning with AL methods without any other enhancements) may be less effective when used with LMs due to their inconsistency created by selecting harmful unlearnable outliers or samples that create instability in the optimization process \footnote{The definition of instability is taken from \citet{d2022limitations} to be the variance in test set accuracy observed when training multiple different random-seeded models on the same set of data.} \citep{d2022limitations}.

One promising idea is to cluster data based on the quality of their representation. For example, compared to the TF-IDF method \citep{xie2016unsupervised}, BERT showed better results in many metrics \citep{subakti2022performance}. Clustering can also be used for the initialization of AL \citep{nguyen2004active} to propagate the classification decision of the classifier trained on representative samples. Clustering can also be performed during the training process after each iteration as in \citet{hassan2023d}, where the top ten most informative samples of each cluster are used to train the model.


While prior work has explored architectural modifications, data efficiency techniques, and stability enhancements individually, our work provides a comprehensive empirical investigation into their concurrent application and interaction to enhance low-resource  BERT fine-tuning, focusing on the interplay between architectural modifications, data selection efficiency, and fine-tuning stability.

\section{Methodologies}

\subsection{Epistemic neural network}

To improve the architecture, we use ENNs that are particularly well-suited for low-resource sentiment analysis because they explicitly model epistemic uncertainty (and differentiate from aleatoric uncertainty), allowing the active learning process to focus on samples where the model is most unsure. This leads to more efficient exploration of the data space and reduces the risk of overfitting to noisy labels. ENNs $f_{\theta}$ consist of two parts: base model (BERT) $b_{\varsigma}$ and Epinet network $e_{\eta}$. The Epinet, as described in \citet{osband2023epistemic}, consists of two Multi-layer perceptrons with one hidden layer each called prior and learnable. The prior network $e^P$ has no trainable parameters and serves to induce some prior knowledge about uncertainty as a variation of the ENN output \citep{osband2018randomized}. In the learnable network $e^L_{\eta}$, weights are initialized with Glorot initialization \citep{glorot2010understanding} and trained to provide meaningful predictions for all probable epistemic index values $z$. The mentioned process can be described by the equation $f_{\theta}(x, z) = b_{\varsigma}(x) +  e_{\eta}(r_{\varsigma}(x), z)$.


The output of the Epinet $e$ is calculated as the sum of the results of learnable $e^L_{\eta}$ and prior network $e^P$: $e_{\eta}(r(x), z) = e^P(r(x), z) + e^L_{\eta}(r(x), z)$.


The Epinet input is comprised of the features of the base network $r_{\varsigma}(x)$ (in our case the final hidden layer) on an input sample $x$ and an epistemic index $z$ sampled from a standard Gaussian distribution.

\subsection{Active learning}

We focus on three AL methods: entropy, bald and variance sampling \citep{osband2022fine}. These methods called acquisition functions prioritize samples whose labels are most uncertain to the model. For entropy defined as $\mathbb{H}[p] = \sum_xp(x)logp(x)$, the entropy acquisition function is defined as $g^{entropy}(\theta,x) = \mathbb{H}[p(\cdot|x,\theta)]$. The other two methods use an epistemic index to determine uncertainty. Bald acquisition function $ g^{bald}(\theta,x)= \mathbb{H} [p(\cdot | \theta,x)] - \int_{z}P_z(dz)\mathbb{H}[p(\cdot | \theta,x,z)]$ is based on mutual information gain. Variance function $g^{variance}(\theta,x)=\sum_{c}\int_{z}P_z(dz)(p(c|\theta,x,z) - p(c|\theta,x))^2$ uses variation in probabilities.

The model is fine-tuned on a task it has not been trained on. Therefore, a new classification random-weighted layer could make the predictions random at first. This phenomenon, known as cold start, can create a situation where the model is assumed to make non-sensical decisions, since it has not yet been trained on any annotated data \citep{jin2022cold}. To examine this, we fine-tune the model in both cold (classic fine-tuning) and "warm" manner (first epoch uses half of the dataset at random). 

\subsection{Data sampling}

Data sampling in AL can be performed after each training epoch or after each training step \citep{settles2009active}, affecting both the frequency of data acquisition and the computational cost (e.g., loading a subset vs the entire dataset). In experiments, we adopt training-step sampling. Departing from the conventional AL framework, we introduce two distinct fine-tuning approaches: \texttt{Accumulating} and \texttt{Recalculating}. In the \texttt{Accumulating} approach, all data sampled in previous epochs are retained and used for subsequent fine-tuning aligning with standard AL practices. The \texttt{Recalculating} approach re-samples data in each epoch independently, as if each were the initial training iteration. This design gives the model greater flexibility to select informative data at each stage. While \texttt{Accumulating} fine-tuning is widely used, \texttt{Recalculating} fine-tuning represents a novel contribution. Further details on these methods are provided in Appendix \ref{sec:appendix}.

\subsection{Clustering}

To support the findings of \citet{hu2010off}, we apply Agglomerative Hierarchical Clustering \citep{voorhees1986effectiveness} using Ward’s linkage, shown effective in capturing semantic relationships in text embeddings \citep{sharma2019comparative}. Clustering starts with each sample’s embedding as a separate cluster, sequentially merging them bottom-up. We applied clustering in two modes: (1) init clustering, performed before fine-tuning to sample from each cluster, and (2) dynamic clustering, repeated before each epoch similarly to \citet{hassan2023d}.

To test the dependence of acquisition functions on the model itself, we also propose \texttt{Furthest-batch} acquisition function that selects samples furthest from the medoid (sample closest to centroid) of cluster. This function is motivated by the hypothesis that samples located far from the medoid represent boundary cases or diverse viewpoints within the cluster that are not necessarily outliers. By prioritizing these samples, we aim to improve the model's ability to generalize to unseen data. It also enables comparison with uncertainty-based methods and may reveal the extent to which fine-tuning depends on the model vs. dataset.

\subsection{Active learning scheduling}

Similarly to \citet{gonsior2024comparing} but more generalizing, we introduce several Active Learning schedulers that modulate the number of samples used. We hypothesize that this may reduce reliance on harmful outliers, minimize annotation costs, and help prevent issues like performance degradation.

The samples are passed through the model to obtain feature vectors, which are evaluated by the AL method to yield values (\texttt{AL results}) and the corresponding indices indicating sample ranking (\texttt{AL order}). The base scheduler selects 75\% of the data according to \texttt{AL order}, functioning similarly to standard AL. The prob scheduler also selects 75\%, but samples without replacement using normalized \texttt{AL results} as probabilities.

Linear schedulers reduce annotation usage progressively after each epoch. They initially select 50\% of the data for both warm and cold fine-tuning, decreasing to 20\%, 15\%, 10\%, and finally 5\%. Like the base scheduler, they either select deterministically by \texttt{AL order} (linear scheduler) or probabilistically (linear prob scheduler).

Dif-build and dif-build-unique schedulers operate differently. They sort \texttt{AL results}, compute the average pairwise difference, and locate the first index where the next difference exceeds this average. This index sets the sampling threshold. Sampling proceeds as in prob schedulers, either with or without replacement, forming different variants.

\section{Experiments}

We trained the models for 5 epochs for the sentiment analysis task while using the Adam optimizer \citep{kingma2014adam}. All models are based on the BERT architecture and pre-trained in the respective lower-resourced language, namely SlovakBERT \citep{pikuliak2021slovakbert}, IceBERT \citep{snaebjarnarson2022warm}, BERTu \citep{BERTu} and BERTurk models \citep{stefan_schweter_2020_3770924}. All hyperparameters are detailed in the Appendix \ref{sec:appendix}. As baselines, we employ the classic BERT fine-tuning approach, wherein a pre-trained BERT model is further trained on the downstream task with task-specific labels.

The assumption of low-resource setting holds true without modifications in experiments with languages spoken by small amounts of people, such as Maltese. To simulate a low-resource scenario, we intentionally down-sample the dataset for languages such as Turkish, as described in Appendix \ref{sec:appendix}. This enables controlled comparison across languages and aligns with recent approaches. 

\begin{table}[h]
  \centering
  \begin{tabular}{llll}
    \hline
    \textbf{Language}  & \textbf{Size} & \textbf{Split} & \textbf{\#Classes}\\
    \hline
    Slovak     &   30k  &    83-8.5-8.5 &   3   \\
    Maltese    &    851  &    70-10-20 &    2       \\
    Icelandic    &    25k  &    80-10-10 &    2       \\
    Turkish    &    25k  &   80-10-10     &    3  \\    
                                    \hline
  \end{tabular}

  \caption{Specifications of datasets and their splits.}
  \label{tab:datasets}
\end{table}

For evaluation, we used monolingual sentiment analysis datasets detailed in Table \ref{tab:datasets}. In experiments, we fine-tune each model three times and report the average. More details are presented in the Appendix \ref{sec:appendix}.

\section{Results}
\label{sec:results}

The key results of our study are summarized below, with additional details provided in Appendix \ref{sec:appendix}. Table \ref{tab:bestf1} presents a comparison between the base model and the best-performing AL model based on F1 score. Models using Active Learning largely outperform baseline models. We also performed a Student's paired t-test to evaluate the statistical significance of our findings, with a significance level of $p$ < 0.05. All top-performing AL models incorporate both clustering and scheduling strategies.

\begin{table}[h]
  \centering
  \begin{tabular}{lll}
    \hline
    \textbf{Model}  & \textbf{Baseline F1} & \textbf{Best AL F1}\\
    \hline
    SlovakBERT     &   70.98   &    71.83 \\
    BERTu    &    82.08       &    86.10$^{\dag}$ \\
    IceBERT    &    77.89  &    80.06$^{\dag}$ \\
    BERTurk    &    91.06  &   92.16$^{\dag}$ \\    
                                    \hline
  \end{tabular}

  \caption{Results of the best AL models according to the F1 score compared to their respective baselines, with $^{\dag}$~indicating statistical significance.}
  \label{tab:bestf1}
\end{table}

\begin{figure}[t]
\center
  \includegraphics[width=\columnwidth]{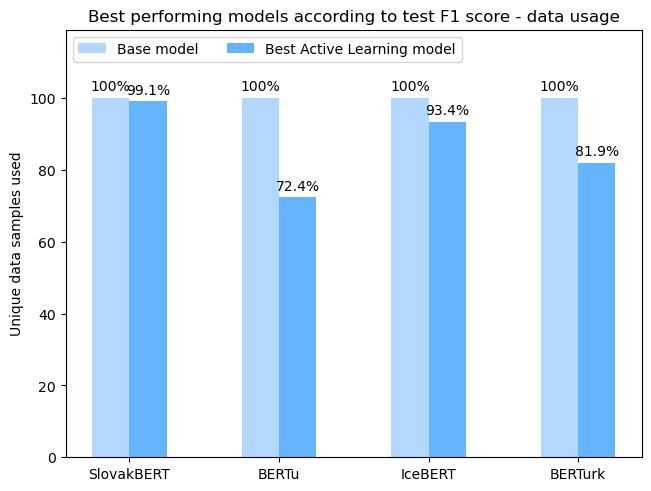} 
  \caption {Data used in the fine-tuning of the models.}
  \label{fig:bestdata}
\end{figure}

\begin{figure}[t]
\center
  \includegraphics[width=\columnwidth]{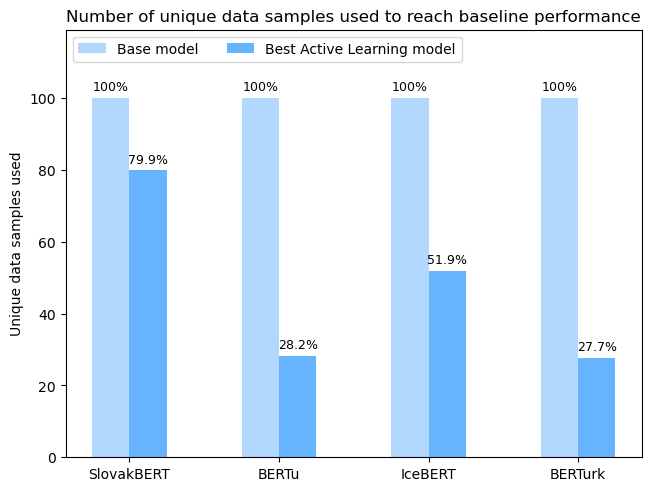} 
  \caption {Data used in the fine-tuning of the models reaching baseline performance while using fewest data samples possible.}
  \label{fig:bestdatabaseline}
\end{figure}

Figure \ref{fig:bestdata} shows the dataset fraction needed to reach the F1 scores in Table \ref{tab:bestf1}, suggesting that data usage often drops markedly even as performance improves. Even more, to reach at least the baseline performance with Active Learning models, amount of data samples needed reduce drastically. We present this fact with Figure \ref{fig:bestdatabaseline}. It is apparent that if the fine-tuning is more focused on saving data annotations rather than enhancing performance (but still keeping it at least comparable), the AL is far superior to the classic fine-tuning process. For more details see Appendix \ref{apx:results}.

\begin{figure}[t] 
\center
  \includegraphics[width=\columnwidth]{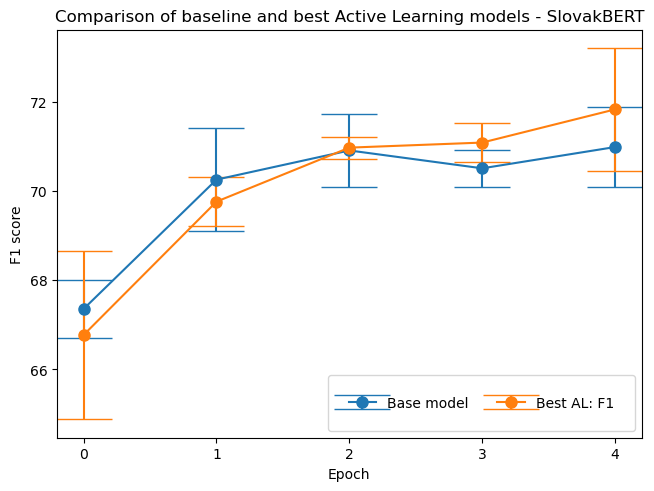} 
  \caption {Learning curves of SlovakBERT models.}
  \label{fig:bestlearningcurve}
\end{figure}

Experiments also indicate that AL contributes to more stable fine-tuning, with fewer instances of performance degradation. This is presented in Figure\ref{fig:bestlearningcurve}, where for the sake of clarity we show the comparison of SlovakBERT base model and best AL model performance comparison.

Our results show that the proposed methods are most effective when combined. To assess individual component contributions in Active Learning, we conducted controlled experiments with the BERTurk model, keeping all settings constant (accumulating sampling, cold start, vanilla BERT architecture, no clustering, no scheduling, and entropy acquisition) while modifying one component at a time. Table \ref{tab:comparechar} presents the best performing configurations based on the F1 score.

\begin{table}[t] %
  \centering
  \begin{tabular}{lll}
    \hline
    \textbf{Characteristic}  & \textbf{F1 score} & \textbf{\% of data} \\
    \hline
    Dynamic clustering  &  91.85   &    96.81\%   \\
    Recalculating sampling    &  91.70   &    81.86\%   \\
    Prob scheduler    &  91.31   &    97.54\%   \\ 
    Bald acquisition &  91.15   &    96.4\%   \\
    ENN architecture &  91.14   &    97.54\%   \\
    Linear scheduler    &  91.10   &    61.49\%   \\
  
  \hline
  \end{tabular}

  \caption{Comparison of the best BERTurk results of different fine-tuning settings.}
  \label{tab:comparechar}
\end{table}

\begin{figure}[t] 
\center
  \includegraphics[width=\columnwidth]{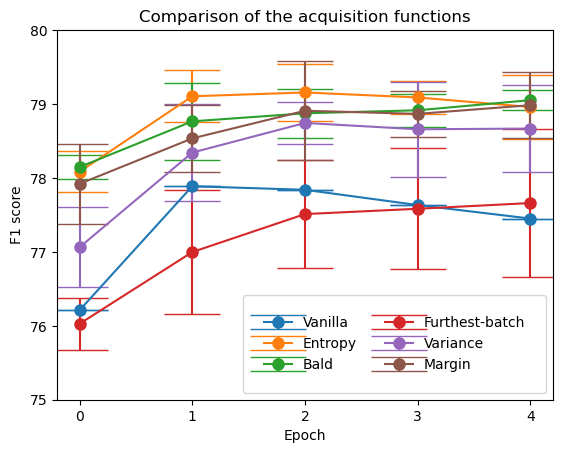} 
  \caption {Learning curves of IceBERT models with different acquistion functions.}
  \label{fig:bestaqs}
\end{figure}

Finally, we also present a comparison of the acquisitions functions. In Figure \ref{fig:bestaqs}, we show a comparison of the best IceBERT models using different acquisition functions. Results indicate that using any acquisition function apart from Furthest-batch is beneficial to the fine-tuning while differences among them are very small. Furthest-batch seems to have similar performance as the baseline model but with much less data used (see Appendix \ref{apx:results} for more information). Interestingly, while fine-tuning for more epochs seems to be crucial for most of the acquistion function to enhance the performance to the maximum, baseline model started to slightly decline after just 2 epochs.

To evaluate parameter-efficient strategies, we conducted controlled experiments using LoRA adapters \citep{hu2022lora}, keeping data splits and training sizes identical to the full fine-tuning baseline. We explored multiple configurations by varying LoRA rank, learning rate, and dropout.
Across all languages and setups, LoRA consistently underperformed full fine-tuning, and performance degraded substantially on the low-resource Maltese condition.
These results contrast with prior work reporting strong LoRA performance on large-scale English benchmarks, indicating that the parameter-efficient updates were not expressive enough to model the language-specific sentiment patterns required in our experiments.

\begin{table}[t]
\centering
\begin{tabular}{lll}
\hline
\textbf{Language} & \textbf{Best AL} & \textbf{Best LoRA} \\
\hline
Slovak & +1.20\% & -5.11\% \\
Maltese & +4.90\% & -51.00\%  \\
Icelandic & +2.79\% & -0.60\%  \\
Turkish & +1.21\% & -0.96\% \\
\hline
\end{tabular}
\caption{Relative performance changes in F1 score using Active Learning or LoRA when compared to the respective baseline models.}
\label{tab:gainlossall}
\end{table}

\section{Conclusions}

Our findings demonstrate that integrating Active Learning with clustering and scheduling yields significant improvements for low-resource language models. This systematic approach reduced annotation requirements by up to 27.6\% for IceBERT and 18.1\% for BERTurk while simultaneously improving performance, with statistically significant gains in the F1 score of up to 4.02 points for BERTu and 2.17 points for IceBERT. Furthermore, combined methodology notably improved fine-tuning stability, reducing performance fluctuations across training epochs. While effectiveness varies by language and dataset characteristics, empirical evidence consistently supports the value of our approach, possibly generalizing to medium and high-resource languages such as Turkish, where particularly the recalculating sampling strategy and linear scheduler achieved a strong result with 81.9\% and 61.49\% of training data, respectively. Moreover, the proposed fine-tuning pipeline can be easily extended to other classification tasks, such as Named Entity Recognition, where only slight changes to architecture and acquisition functions are needed.

\section{Limitations}

\subsection{Datasets}

The datasets used in this work were expected to be reasonably small, either by choice (e.g., using a portion of a large dataset) or by necessity. It is understandable that the size of the dataset influences expected savings on annotations. In this work, 2 of the datasets can already be considered small. From two of the datasets with a reasonable amount of data, only a smart portion was used to show the possible enhancements presented in this work.

\subsection{Models}

In comparison to state-of-the-art architectures, the BERT and similar models used in our experiments are only a fraction of their size. Nevertheless, for fine-tuning, low-resource languages do not have many other choices. As BERT models are still relatively popular, the decision to use them in our work was made, although we acknowledge that for some (or most) use cases, using bigger decoder-style models can obtain better performance.

\subsection{Experiments}

For each combination of model's important fine-tuning characteristic (AL method, AL scheduling, clustering, ...), we conducted three runs and calculated the average that represents the final result. While this aims to eliminate model bias, a larger number of runs might help reduce it much more.

For Turkish and Icelandic, we simulate low-resource conditions via controlled downsampling, which, while enabling fair comparison, may not fully capture the challenges of genuine low-resource environments.

Due to computational constraints, we were unable to explore all possible combinations of AL methods, clustering techniques, and scheduling strategies. Future work should investigate a wider range of configurations, particularly those excluded from this study as mentioned in Appendix \ref{sec:appendix_exp}.

\subsection{Results}

We argue that the presented results show that using Active Learning with clustering and scheduling enhances the fine-tuning process of lower-resourced language BERT models while also enhancing F1 scores. Further experiments on more languages could strenghten this claim; nevertheless, our results on bigger and more used language (Turkish) can serve as a form of generalization on also medium-resource languages.

The differences observed across experimental setups are likely attributable to the specific characteristics of each language model and dataset. While our results demonstrate that Active Learning and clustering generally lead to improved performance, we acknowledge that tailoring acquisition functions and scheduling strategies to individual model–dataset combinations can yield even greater gains. Therefore, it is highly plausible that conducting targeted experiments with varied acquisition functions and scheduling approaches is always beneficial for identifying the most effective configuration in each specific context.

\subsection*{Acknowledgements}

This work was supported by grant APVV-21-0114.

\bibliography{custom}

@article{devlin2018bert,
  title={Bert: Pre-training of deep bidirectional transformers for language understanding},
  author={Devlin, Jacob},
  journal={arXiv preprint arXiv:1810.04805},
  year={2018}
}

@article{achiam2023gpt,
  title={Gpt-4 technical report},
  author={Achiam, Josh and Adler, Steven and Agarwal, Sandhini and Ahmad, Lama and Akkaya, Ilge and Aleman, Florencia Leoni and Almeida, Diogo and Altenschmidt, Janko and Altman, Sam and Anadkat, Shyamal and others},
  journal={arXiv preprint arXiv:2303.08774},
  year={2023}
}

@article{osband2023epistemic,
  title={Epistemic neural networks},
  author={Osband, Ian and Wen, Zheng and Asghari, Seyed Mohammad and Dwaracherla, Vikranth and Ibrahimi, Morteza and Lu, Xiuyuan and Van Roy, Benjamin},
  journal={Advances in Neural Information Processing Systems},
  volume={36},
  pages={2795--2823},
  year={2023}
}

@incollection{short2018czech,
  title={Czech and slovak},
  author={Short, David},
  booktitle={The world's major languages},
  pages={314--338},
  year={2018},
  publisher={Routledge}
}

@article{al2017aroma,
  title={Aroma: A recursive deep learning model for opinion mining in arabic as a low resource language},
  author={Al-Sallab, Ahmad and Baly, Ramy and Hajj, Hazem and Shaban, Khaled Bashir and El-Hajj, Wassim and Badaro, Gilbert},
  journal={ACM Transactions on Asian and Low-Resource Language Information Processing (TALLIP)},
  volume={16},
  number={4},
  pages={1--20},
  year={2017},
  publisher={ACM New York, NY, USA}
}

@article{bojanowski2017enriching,
  title={Enriching word vectors with subword information},
  author={Bojanowski, Piotr and Grave, Edouard and Joulin, Armand and Mikolov, Tomas},
  journal={Transactions of the association for computational linguistics},
  volume={5},
  pages={135--146},
  year={2017},
  publisher={MIT Press One Rogers Street, Cambridge, MA 02142-1209, USA journals-info~…}
}

@article{heinzerling2017bpemb,
  title={BPEmb: Tokenization-free pre-trained subword embeddings in 275 languages},
  author={Heinzerling, Benjamin and Strube, Michael},
  journal={arXiv preprint arXiv:1710.02187},
  year={2017}
}

@article{settles2009active,
  title={Active learning literature survey},
  author={Settles, Burr},
  year={2009},
  publisher={University of Wisconsin-Madison Department of Computer Sciences}
}

@inproceedings{lewis1995sequential,
  title={A sequential algorithm for training text classifiers: Corrigendum and additional data},
  author={Lewis, David D},
  booktitle={Acm Sigir Forum},
  volume={29},

  pages={13--19},
  year={1995},
  organization={ACM New York, NY, USA}
}

@article{shannon1948mathematical,
  title={A mathematical theory of communication},
  author={Shannon, Claude Elwood},
  journal={The Bell system technical journal},
  volume={27},
  number={3},
  pages={379--423},
  year={1948},
  publisher={Nokia Bell Labs}
}

@article{d2022limitations,
  title={Limitations of active learning with deep transformer language models},
  author={D'Arcy, Mike and Downey, Doug},
  year={2022}
}

@article{osband2022fine,
  title={Fine-tuning language models via epistemic neural networks},
  author={Osband, Ian and Asghari, Seyed Mohammad and Van Roy, Benjamin and McAleese, Nat and Aslanides, John and Irving, Geoffrey},
  journal={arXiv preprint arXiv:2211.01568},
  year={2022}
}

@inproceedings{glorot2010understanding,
  title={Understanding the difficulty of training deep feedforward neural networks},
  author={Glorot, Xavier and Bengio, Yoshua},
  booktitle={Proceedings of the thirteenth international conference on artificial intelligence and statistics},
  pages={249--256},
  year={2010},
  organization={JMLR Workshop and Conference Proceedings}
}

@article{osband2018randomized,
  title={Randomized prior functions for deep reinforcement learning},
  author={Osband, Ian and Aslanides, John and Cassirer, Albin},
  journal={Advances in Neural Information Processing Systems},
  volume={31},
  year={2018}
}

@article{subakti2022performance,
  title={The performance of BERT as data representation of text clustering},
  author={Subakti, Alvin and Murfi, Hendri and Hariadi, Nora},
  journal={Journal of big Data},
  volume={9},
  number={1},
  pages={15},
  year={2022},
  publisher={Springer}
}

@inproceedings{xie2016unsupervised,
  title={Unsupervised deep embedding for clustering analysis},
  author={Xie, Junyuan and Girshick, Ross and Farhadi, Ali},
  booktitle={International conference on machine learning},
  pages={478--487},
  year={2016},
  organization={PMLR}
}

@inproceedings{nguyen2004active,
  title={Active learning using pre-clustering},
  author={Nguyen, Hieu T and Smeulders, Arnold},
  booktitle={Proceedings of the twenty-first international conference on Machine learning},
  pages={79},
  year={2004}
}

@article{hassan2023d,
  title={D-CALM: A dynamic clustering-based active learning approach for mitigating bias},
  author={Hassan, Sabit and Alikhani, Malihe},
  journal={arXiv preprint arXiv:2305.17013},
  year={2023}
}

@inproceedings{hu2010off,
  title={Off to a good start: Using clustering to select the initial training set in active learning},
  author={Hu, Rong and Mac Namee, Brian and Delany, Sarah Jane},
  booktitle={Twenty-Third International FLAIRS Conference},
  year={2010}
}

@book{voorhees1986effectiveness,
  title={The effectiveness and efficiency of agglomerative hierarchic clustering in document retrieval},
  author={Voorhees, Ellen Marie},
  year={1986},
  publisher={Cornell University}
}

@inproceedings{sharma2019comparative,
  title={Comparative study of single linkage, complete linkage, and ward method of agglomerative clustering},
  author={Sharma, Shweta and Batra, Neha and others},
  booktitle={2019 international conference on machine learning, big data, cloud and parallel computing (COMITCon)},
  pages={568--573},
  year={2019},
  organization={IEEE}
}

@article{kingma2014adam,
  title={Adam: A method for stochastic optimization},
  author={Kingma, Diederik P},
  journal={arXiv preprint arXiv:1412.6980},
  year={2014}
}

@article{pikuliak2021slovakbert,
  title={SlovakBERT: Slovak masked language model},
  author={Pikuliak, Mat{\'u}{\v{s}} and Grivalsk{\`y}, {\v{S}}tefan and Kon{\^o}pka, Martin and Bl{\v{s}}t{\'a}k, Miroslav and Tamajka, Martin and Bachrat{\`y}, Viktor and {\v{S}}imko, Mari{\'a}n and Bal{\'a}{\v{z}}ik, Pavol and Trnka, Michal and Uhl{\'a}rik, Filip},
  journal={arXiv preprint arXiv:2109.15254},
  year={2021}
}

@inproceedings{BERTu,
    title = "Pre-training Data Quality and Quantity for a Low-Resource Language: New Corpus and {BERT} Models for {M}altese",
    author = "Micallef, Kurt  and
              Gatt, Albert  and
              Tanti, Marc  and
              van der Plas, Lonneke  and
              Borg, Claudia",
    booktitle = "Proceedings of the Third Workshop on Deep Learning for Low-Resource Natural Language Processing",
    month = jul,
    year = "2022",
    address = "Hybrid",
    publisher = "Association for Computational Linguistics",
    url = "https://aclanthology.org/2022.deeplo-1.10",
    doi = "10.18653/v1/2022.deeplo-1.10",
    pages = "90--101",
}

@software{stefan_schweter_2020_3770924,
  author       = {Stefan Schweter},
  title        = {BERTurk - BERT models for Turkish},
  month        = apr,
  year         = 2020,
  publisher    = {Zenodo},
  version      = {1.0.0},
  doi          = {10.5281/zenodo.3770924},
  url          = {https://doi.org/10.5281/zenodo.3770924}
}

@inproceedings{dingli2016sentiment,
  title={Sentiment analysis on Maltese using machine learning},
  author={Dingli, Alexiei and Sant, Nicole},
  booktitle={Proceedings of The Tenth International Conference on Advances in Semantic Processing (SEMAPRO 2016)},
  pages={21--25},
  year={2016}
}

@inproceedings{cortis-davis-2019-social,
    title = "A Social Opinion Gold Standard for the {M}alta Government Budget 2018",
    author = "Cortis, Keith  and
      Davis, Brian",
    editor = "Xu, Wei  and
      Ritter, Alan  and
      Baldwin, Tim  and
      Rahimi, Afshin",
    booktitle = "Proceedings of the 5th Workshop on Noisy User-generated Text (W-NUT 2019)",
    month = nov,
    year = "2019",
    address = "Hong Kong, China",
    publisher = "Association for Computational Linguistics",
    url = "https://aclanthology.org/D19-5547",
    doi = "10.18653/v1/D19-5547",
    pages = "364--369",
}

@article{vstefanik2023resources,
  title={Resources and Few-shot Learners for In-context Learning in Slavic Languages},
  author={{\v{S}}tef{\'a}nik, Michal and Kadl{\v{c}}{\'\i}k, Marek and Gramacki, Piotr and Sojka, Petr},
  journal={arXiv preprint arXiv:2304.01922},
  year={2023}
}

@article{snaebjarnarson2022warm,
  title={A Warm Start and a Clean Crawled Corpus--A Recipe for Good Language Models},
  author={Sn{\ae}bjarnarson, V{\'e}steinn and S{\'\i}monarson, Haukur Barri and Ragnarsson, P{\'e}tur Orri and Ing{\'o}lfsd{\'o}ttir, Svanhv{\'\i}t Lilja and J{\'o}nsson, Haukur P{\'a}ll and {\TH}orsteinsson, Vilhj{\'a}lmur and Einarsson, Hafsteinn},
  journal={arXiv preprint arXiv:2201.05601},
  year={2022}
}

@article{nigatu2024zeno,
  title={The Zeno's Paradox ofLow-Resource'Languages},
  author={Nigatu, Hellina Hailu and Tonja, Atnafu Lambebo and Rosman, Benjamin and Solorio, Thamar and Choudhury, Monojit},
  journal={arXiv preprint arXiv:2410.20817},
  year={2024}
}

@article{jukic2023parameter,
  title={Parameter-Efficient Language Model Tuning with Active Learning in Low-Resource Settings},
  author={Juki{\'c}, Josip and {\v{S}}najder, Jan},
  journal={arXiv preprint arXiv:2305.14576},
  year={2023}
}

@inproceedings{wertz2023reinforced,
  title={Reinforced active learning for low-resource, domain-specific, multi-label text classification},
  author={Wertz, Lukas and Bogojeska, Jasmina and Mirylenka, Katsiaryna and Kuhn, Jonas},
  booktitle={Findings of the Association for Computational Linguistics: ACL 2023},
  pages={10959--10977},
  year={2023}
}

@article{jin2022cold,
  title={Cold-start active learning for image classification},
  author={Jin, Qiuye and Yuan, Mingzhi and Li, Shiman and Wang, Haoran and Wang, Manning and Song, Zhijian},
  journal={Information sciences},
  volume={616},
  pages={16--36},
  year={2022},
  publisher={Elsevier}
}

@article{shin2023prompt,
  title={Prompt engineering or fine tuning: An empirical assessment of large language models in automated software engineering tasks},
  author={Shin, Jiho and Tang, Clark and Mohati, Tahmineh and Nayebi, Maleknaz and Wang, Song and Hemmati, Hadi},
  journal={arXiv preprint arXiv:2310.10508},
  year={2023}
}

@article{gonsior2024comparing,
  title={Comparing and Improving Active Learning Uncertainty Measures for Transformer Models by Discarding Outliers},
  author={Gonsior, Julius and Falkenberg, Christian and Magino, Silvio and Reusch, Anja and Hartmann, Claudio and Thiele, Maik and Lehner, Wolfgang},
  journal={Information systems frontiers},
  pages={1--17},
  year={2024},
  publisher={Springer}
}

@article{hu2022lora,
  title={Lora: Low-rank adaptation of large language models.},
  author={Hu, Edward J and Shen, Yelong and Wallis, Phillip and Allen-Zhu, Zeyuan and Li, Yuanzhi and Wang, Shean and Wang, Lu and Chen, Weizhu and others},
  journal={ICLR},
  volume={1},
  number={2},
  pages={3},
  year={2022}
}

\onecolumn
\clearpage
\appendix

\section{Appendix}
\label{sec:appendix}

\subsection{Datasets}

The largest dataset used  was the Slovak ČSFD dataset \citep{vstefanik2023resources} with 30k annotated samples containing movie reviews.  Although the original dataset includes six sentiment labels, we simplify the classification task to three classes by applying the following mapping: [0, 1 → 0], [2, 3 → 1], and [4, 5 → 2]. 

While Turkish is not a low-resource language, we intentionally downsampled the dataset comprising tweets and other sentiment-related text samples to 5\% to investigate the performance of active learning methods in extremely data-scarce scenarios. This allowed us to evaluate the robustness of our approach when annotation budgets are severely limited.

For the Maltese language, we use the joint datasets of \citep{cortis-davis-2019-social} and \citep{dingli2016sentiment} containing data originating from comments on news articles and social media posts. Similarly as in Turkish dataset, in Icelandic dataset, which was created by translating IMDB movie reviews, we down-sample to 25k annotated data samples to mimic low-resource scenario \footnote{All datasets are available on Huggingface website, namely Slovak sk\_csfd-movie-reviews, Turkish turkish-sentiment-analysis-dataset,  Maltese maltese\_sa and and Icelandic imdb-isl-mideind-translate.}. Selection of the relatively small and specialized datasets stems from the setup of our research task, which tries to mimic many real-life scenarios (particularly in the context of lower-resourced languages), in which only a small task-specific dataset is available and/or where expert annotation is very costly. This in turn creates the need to make maximal use of the available data. Additional details about datasets are presented in Table \ref{tab:datasets_add}.

\begin{table}[h]
  \centering
  \begin{tabular}{lllll}
    \hline
    \textbf{Dataset}  & \textbf{Class 1} & \textbf{Class 2} & \textbf{Class 3} & \textbf{Avg token number}\\
    \hline
    Slovak     &   33.2\%  &    33.4\% &   33.4\%  & 368  \\
    Maltese    &   60\%  &    40\% &   -  & 123  \\
    Icelandic  &   50\%  &    50\% &  -  &   1367 \\
    Turkish    &   53.5\%  &    34.9\% &   11.6\%  & 140  \\    
                                    \hline
  \end{tabular}

  \caption{Specifications of datasets' class distribition and average token length.}
  \label{tab:datasets_add}
\end{table}

\subsection{Models}

Some of the most important specifications of the models used can be seen in Table \ref{table:low-res-compar}:

\begin{table}[h]
\centering{%
\begin{tabular}{ |c|c|c|c|c| } 
\hline
Characteristic$\setminus$Model & SlovakBERT & IceBERT & BERTurk & BERTu\\
\hline
Architecture & RoBERTa & RoBERTa & BERT & BERT\\
\hline
\#Layers & 12 & 12 & 12 & 12 \\
\#Heads & 12 & 12 & 12 & 12\\
Hidden size &  768 & 768 & 768  & 768\\
\#Parameters & 125M & 125M & 109M  & 560M\\
\#Training tokens & 4.6B & 2.7B & 44.04B & 500M  \\
\#Training steps & 300K & 225K & 2M  & 1M  \\
Vocabulary size & 50K & 50K & 32K &  52K \\
Pretraining data size & 19.35GB & 15.8GB & 35GB &  2.52GB \\
\hline
\end{tabular}}
\caption{Important specifications of models used in experiments.}
\label{table:low-res-compar}
\end{table}

There are two main reasons why we use BERT-based models:

\begin{itemize}
    \item For many lower-resourced languages, BERT-based models are the best (and often the only) available Pre-trained Language Model option.
    \item The model family is fairly easy to use, as well as resource-efficient, and as such has a higher chance to reach a wider research audience.
\end{itemize}

\subsection{Experiments}
\label{sec:appendix_exp}

Table \ref{tab:parameters} presents important parameters for our low-setting environment:

\begin{table}[h]
  \centering
  \begin{tabular}{ll}
    \hline
    \textbf{Parameter}  & \textbf{Value} \\
    \hline
    Epochs     &   5   \\
    Max text length    &    64             \\
    Batch size    &    32          \\   
    Learning rate    &    $5 * 10^{-5}$          \\ 
                                    \hline
  \end{tabular}

  \caption{Specifications of parameters used in experiments.}
  \label{tab:parameters}
\end{table}

If we assume that at the start of the training (in our case fine-tuning) we have an unlabeled pool $U_0$ with all data samples from the dataset and an empty annotated data pool $D_0$, the difference in one training epoch between Accumulating and Recalculating fine-tuning can be summarized as:

\paragraph{Accumulating (building) fine-tuning}
\begin{enumerate} 
\item Load data samples from previous epoch unlabeled data pool $U_{e-1}$ and used annotated data pool $D_{e-1}$.
\item Sample up to 75\% new data samples from $U_{e-1}$, annotate them and add them to $D_{e-1}$, thus creating new data pools $U_{e}$ (without newly sampled data) and $D_{e}$ (with newly sampled data).
\item Fine-tune the model with all the accumulated data in the dataset $D_{e}$.
\end{enumerate}

\paragraph{Recalculating fine-tuning}
\begin{enumerate} 
\item Sample up to 75\% of the data samples from $U_{0}$, annotate them and add them to empty $D_{0}$, while retaining the same data pool $U_{0}$ and creating the data pool $D_{e}$ (with sampled data).
\item Fine-tune the model with all the data in the dataset $D_{e}$.
\item Delete the whole $D_{e}$.
\end{enumerate}

As the model's weights are changing and the model is learning, it can decide that some of the data samples used previously can be no longer helpful or even misleading (e.g. outliers). This fact is not relevant in Accumulating fine-tuning, which uses these samples nevertheless, but serves as a motivation for Recalculating fine-tuning. One important fact of the Recalculating fine-tuning: since this happens in multiple epochs, some data can be chosen multiple times (and some will not be chosen at all); still, most of the dataset or the whole dataset may be used (more precisely: the data sample can be chosen multiple times, but it will be used at most once during each training epoch since the annotated dataset is built from fresh every epoch.).

Some of the combinations of acquisition function, architecture, clustering, and scheduling are not possible or helpful due to different circumstances. To save on computational resources, their exploration was not conducted. In Tables \ref{tab:possibles}, \ref{tab:possibles_two} and \ref{tab:possibles_three}, we show which of the combinations were not present in our experimental setups:

\begin{table}[h]
\centering
\begin{tabular}{|c|c|c|c|}
\hline
Option  & No clustering & Init clustering & Dynamic clustering \\
\hline
Base / prob scheduler & \cmark    & \cmark  & \cmark  \\
\hline
Linear schedulers &    \cmark  & \warn  & \cmark  \\
\hline
Dif-build schedulers & \xmark  &  \cmark &  \cmark\\
\hline
\end{tabular}
\caption{Setups of different clusterings and schedulers used in the experiments.}
\label{tab:possibles}
\end{table}

\begin{table}[h]
\centering
\begin{tabular}{|c|c|c|c|}
\hline
Option  & Classic AF & Epistemic AF & Furthest-batch AF  \\
\hline
Accumulating base BERT &   \cmark   &     \xmark  & \cmark   \\
\hline
Accumulating ENN &   \cmark  &      \cmark       & \cmark \\
\hline
Recalculating base BERT &  \cmark &   \xmark   & \cmark \\
\hline
Recalculating ENN & \cmark  &    \cmark      & \cmark \\
\hline
\end{tabular}
\caption{Setups of different acquisition functions and architecture types used in the experiments.}
\label{tab:possibles_two}
\end{table}

\begin{table}[h]
\centering
\begin{tabular}{|c|c|c|c|}
\hline
Option  & Classic AF & Epistemic AF & Furthest-batch AF  \\
\hline
Base  scheduler &  \cmark   &  \cmark  &      \cmark     \\
\hline
Prob scheduler &   \cmark  &  \cmark  &        \xmark    \\
\hline
Linear scheduler &  \cmark   &  \cmark  &      \cmark     \\
\hline
Linear prob scheduler &  \cmark    & \cmark  &      \xmark      \\
\hline
Dif-build schedulers &  \cmark  &  \cmark  &    \xmark    \\
\hline
\end{tabular}
\caption{Setups of different acquisition functions and schedulers used in the experiments.}
\label{tab:possibles_three}
\end{table}

Due to the design of dif-build schedulers, it is appropriate to evaluate their performance only on data drawn from the same clusters, rather than from random samples across the entire dataset. Init clustering plays a role solely during the first epoch of fine-tuning. Consequently, combining it with linear schedulers—which adjust the volume of sampled data across epochs—offers no advantage over dynamic clustering (marked as \warn).

Epistemic acquisition functions inherently rely on the epistemic index and the broader Epistemic neural network architecture. In contrast, the furthest-batch acquisition function is based exclusively on data sample characteristics. As a result, model-dependent schedulers such as prob, linear prob and dif-build are not applicable when using furthest-batch. Additionally, because furthest-batch determines acquisition based on distances from medoids, it requires clustering and cannot operate without it.

Although combining furthest-batch with dynamic clustering is feasible, it does not offer any benefit over using it with init clustering. This is because the selected samples remain  unchanged across epochs, rendering dynamic updates unnecessary in this case.

\subsection{Results}
\label{apx:results}

This section presents supplementary results not included in the main text. In Section \ref{sec:results}, we compared the best-performing models against their respective baselines and analyzed the individual contributions of key components of Active Learning (AL). Here, we extend this analysis by evaluating how different aspects of AL, when combined, influence fine-tuning performance—reflecting more realistic usage scenarios.

Specifically, we report the average F1 score gains or losses of models incorporating particular AL characteristics (in combination with others), relative to their corresponding baselines. These comparisons are aggregated across all models and languages, along with the percentage of labeled data utilized. After that, we report individual statistics as in Table \ref{tab:comparechar} for other languages (Slovak, Icelandic, Maltese).

Subsequently, we present the learning curves for BERTu, IceBERT, and BERTurk models to illustrate performance trends throughout the fine-tuning process. Finally, to facilitate direct comparisons of various fine-tuning configurations, we include a series of tables showing the best results per language for the following model variants: (i) the vanilla base model, (ii) the model with only Active Learning, (iii) Active Learning with clustering, and (iv) Active Learning with both clustering and scheduling.

\begin{table}[h]
  \centering
  \begin{tabular}{lll}
    \hline
    \textbf{Characteristic}  & \textbf{+- \% of F1 score} & \textbf{\% of data} \\
    \hline
    \multicolumn{3}{c}{Architecture} \\
    Vanilla     &  +1.89\%   &    90.72\%   \\
    ENN     &  +2.39\%   &    87.59\%   \\
    
    \multicolumn{3}{c}{Start} \\
    Warm     &  +1.74\%   &    93.06\%   \\
    Cold     &  +2.44\%   &    89.12\%   \\
    
    \multicolumn{3}{c}{Sampling} \\
    Recalculating     &  +2.26\%   &    86.43\%   \\
    Accumulating     &  +2.39\%   &    92.54\%   \\

    \multicolumn{3}{c}{Acquisition function} \\
    None   &  -2.67\%    &  99.48\%  \\
    Entropy     &  +2.28\%   &    92.07\%   \\
    Bald     &   +2.32\%  &    87.89\%   \\
    Variance     &   +2.26\%  &    89.72\%   \\   
    Furthest-batch  &  +1.68\%  &  79.81\%   \\

    \multicolumn{3}{c}{Clustering} \\
    No clustering    &  +1.76\%   &    95.51\%   \\
    Init clustering     &   +2.26\%   &    78.66\%   \\
    Dynamic clustering     &   +2.21\%  &    97.12\%   \\

    \multicolumn{3}{c}{Schedulers} \\
    Base &  +2.02\%   &   85.67\%        \\  
    Prob scheduler  &  +2.24\%  &     92.76\%   \\
    Linear scheduler    &  +2.08\%   &   80.68\%   \\
    Linear prob scheduler    &  +1.66\%   &    81.68\%   \\
    Dif-build    &   +1.59\%  &    90.25\%   \\
    Dif-build-unique    &  +1.95\%   &    90.10\%   \\
  
  \hline
  \end{tabular}

  \caption{Comparison of the average results of different fine-tuning settings.}
  \label{tab:compareallchar}
\end{table}

In Table \ref{tab:compareallchar} we present the results of an ablation study designed to compare the impact of various fine-tuning strategies, including different schedulers, clustering methods, and other design choices. Notably, the results highlight the significant role of Active Learning. All tested acquisition functions outperformed the baseline condition without AL, underscoring its overall effectiveness. Among them, the Bald acquisition function achieved the highest F1 scores, while the furthest-batch strategy led to substantial reductions in the number of required annotations.

Fine-tuning BERT models as Epistemic neural networks contributed to both improved performance and reduced data requirements. Similarly, cold fine-tuning yielded beneficial effects, challenging the assumption that possible random fine-tuning during the initial epoch degrades performance. Additionally, recalculating sampling showed comparable performance to traditional accumulation-based sampling, while offering further annotation savings.

Finally, the use of linear schedulers—particularly the standard linear scheduler—proved effective in balancing annotation efficiency with strong model performance.

\begin{table}[h] 
  \centering
  \begin{tabular}{lll}
    \hline
    \textbf{Characteristic}  & \textbf{F1 score} & \textbf{\% of data} \\
    \hline
    ENN architecture &  71.43  &    99.52\%   \\
    Dynamic clustering  &  71.24   &    99.62\%   \\  
    Recalculating sampling    &  71.17   &    94.2\%   \\
  \hline
  \end{tabular}

  \caption{Comparison of the best SlovakBERT results of different fine-tuning settings.}

\end{table}

\begin{table}[h] 
  \centering
  \begin{tabular}{lll}
    \hline
    \textbf{Characteristic}  & \textbf{F1 score} & \textbf{\% of data} \\
    \hline
    Dynamic clustering  &  85.38   &    98.9\%   \\  
    Init clustering &  84.52   &    77.98\%   \\
    Entropy acquisition &  84.35   &    96.02\%   \\
  
  \hline
  \end{tabular}

  \caption{Comparison of the best BERTu results of different fine-tuning settings.}

\end{table}

\begin{table}[h] 
  \centering
  \begin{tabular}{lll}
    \hline
    \textbf{Characteristic}  & \textbf{F1 score} & \textbf{\% of data} \\
    \hline
    Bald acquisition &  79.77   &    93.37\%   \\
    ENN architecture &  79.22  &    93.37\%   \\
    Prob scheduler &  78.23  &    99.5\%   \\  
  
  \hline
  \end{tabular}

  \caption{Comparison of the best ICEBert results of different fine-tuning settings.}

\end{table}


\begin{figure*}[hbt!]
\center
  \includegraphics[width=0.7\linewidth]{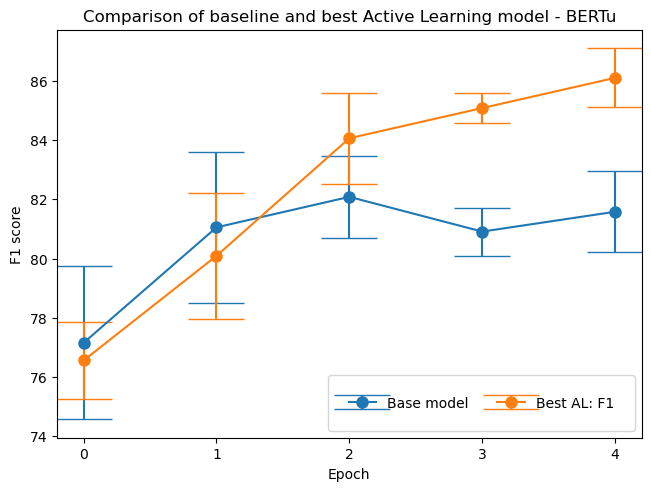} 
  \caption {Learning curves of baseline model and the best performing AL models while fine-tuning BERTu.}
\end{figure*}

\begin{figure*}[hbt!]
\center
  \includegraphics[width=0.7\linewidth]{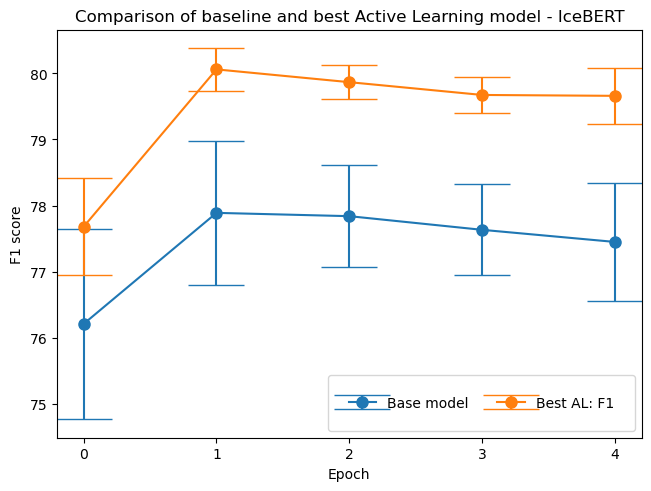} 
  \caption {Learning curves of baseline model and the best performing AL models while fine-tuning IceBERT.}
\end{figure*}

\begin{figure*}[hbt!]
\center
  \includegraphics[width=0.7\linewidth]{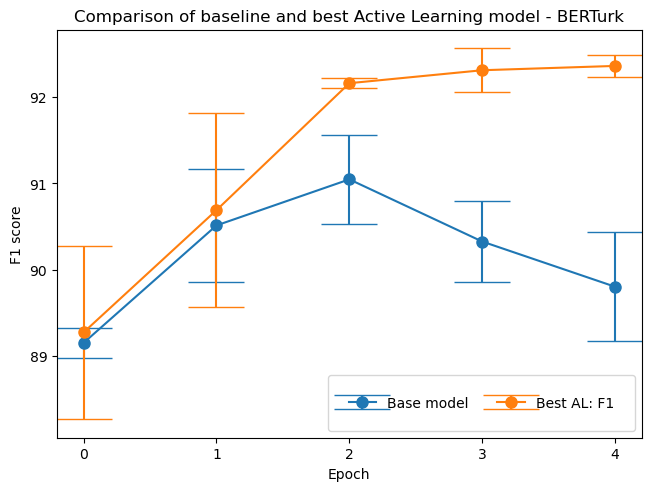} 
  \caption {Learning curves of baseline model and the best performing AL models while fine-tuning BERTurk.}
\end{figure*}


\begin{table}[h]   
  \centering
  \begin{tabular}{lll}
    \hline
    \textbf{Model}  & \textbf{F1 score} & \textbf{\% of data} \\
        \hline
    Vanilla     &  71.17  &    100\%   \\
    Entropy     &  71.23  &    99.92\%   \\  
    Bald + dynamic     &  71.64    &    99.6\%   \\
    Entropy + dynamic + linear prob     &  71.83   &    99.1\%   \\  
  \hline
  \end{tabular}

  \caption{Comparison of the best SlovakBERT models.}
  \label{tab:comparecharslovak}
\end{table}

\begin{table}[h]   
  \centering
  \begin{tabular}{lll}
    \hline
    \textbf{Model}  & \textbf{F1 score} & \textbf{\% of data} \\
    \hline
    Vanilla     &  82.76  &    100\%   \\  
    Entropy     &  84.35  &    96.7\%   \\
    Bald + base     &  85.5   &    78\%   \\ 
    Bald + init + linear     &  86.1   &    72.4\%   \\  
  \hline
  \end{tabular}

  \caption{Comparison of the best BERTu models.}
  \label{tab:comparecharbertu}
\end{table}

\begin{table}[h]   
  \centering
  \begin{tabular}{lll}
    \hline
    \textbf{Model}  & \textbf{F1 score} & \textbf{\% of data} \\
    \hline
    Vanilla     &  78.07  &    100\%   \\
    Bald     &  79.77  &    93.4\%   \\
    Bald + init     &  78.89   &    85.3\%   \\ 
    Entropy + dynamic + prob     &  80.06   &    93.4\%   \\  
  \hline
  \end{tabular}

  \caption{Comparison of the best IceBERT models.}
  \label{tab:comparecharice}
\end{table}

\begin{table}[h]   
  \centering
  \begin{tabular}{lll}
    \hline
    \textbf{Model}  & \textbf{F1 score} & \textbf{\% of data} \\
    \hline
    Vanilla     &  91.06  &    100\%   \\
    Entropy     &  91.3  &    81.9\%   \\  
    Bald + init     &   91.94  &    70.13\%   \\
    Bald + dynamic + prob     &  92.16   &    81.9\%   \\   
  \hline
  \end{tabular}

  \caption{Comparison of the best BERTurk models.}
  \label{tab:comparecharturk}
\end{table}

\end{document}